%% file: main.tex
\documentclass[runningheads]{llncs}

\usepackage[mobile]{eccv}

\input{preamble}

\usepackage{graphicx}
\usepackage{booktabs}

\usepackage[accsupp]{axessibility}  % Improves PDF readability for those with disabilities.

\usepackage{hyperref}

\usepackage{orcidlink}

\begin{document}

% ---------------------------------------------------------------
% TODO REVIEW: Replace with your title
\title{Generating Multi-view Adversarial Examples for Visual Geometry Grounded Transformer} 

% TODO REVIEW: If the paper title is too long for the running head, you can set
% an abbreviated paper title here. If not, comment out.
\titlerunning{MVAP-G}

% TODO FINAL: Replace with your author list. 
\author{Qi Song \and
Ziyuan Luo\thanks{Corresponding author. This work was carried out at the Renjie Group, Hong Kong Baptist University.} \and
Haoliang Han \and
Renjie Wan}

% TODO FINAL: Replace with an abbreviated list of authors.
\authorrunning{Q.~Song et al.}

% First names are abbreviated in the running head.
% If there are more than two authors, 'et al.' is used.

% TODO FINAL: Replace with your institution list.
\institute{Department of Computer Science, \\ Hong Kong Baptist University, Hong Kong, China \\
\email{\{qisong, ziyuanluo, haolianghan\}@life.hkbu.edu.hk, \\ renjiewan@hkbu.edu.hk}}

\maketitle

\input{sec/0_abstract}

\input{sec/1_introduction}
\input{sec/2_related_work}
\input{sec/3_method}

\input{sec/4_experiments}

\input{sec/5_conclusion}

\section*{Acknowledgements}
Renjie Group is supported by the National Natural Science Foundation of China under Grant No. 62302415, Guangdong Basic and Applied Basic Research Foundation under Grant No.  2024A1515012822, and the Research Grants Council (RGC) of Hong Kong SAR, China, under a GRF Grant 12203124 and an ECS Grant 22201125.  Ziyuan Luo is supported by a fellowship award from the RGC of Hong Kong SAR, China (Project No. HKBU JRFS2627-2S03).

% ---- Bibliography ----
%
% BibTeX users should specify bibliography style 'splncs04'.
% References will then be sorted and formatted in the correct style.
%
% 

\bibliographystyle{splncs04}
\bibliography{reference}

\end{document}

%% file: sec/0_abstract.tex
\begin{abstract}
The Visual Geometry Grounded Transformer (VGGT) enables unified feed-forward 3D reconstruction from multi-view images. However, deploying such a high-performance model may expose critical security vulnerabilities. Traditional adversarial perturbations require costly per-scene optimization, while Universal Adversarial Perturbations (UAPs) rely on a single static pattern and fail to effectively attack VGGT. To address these limitations, we propose \textbf{MVAP-G}, a multi-view adversarial perturbation generator that produces imperceptible consistent perturbations across multiple views in a single feed-forward pass. To ensure perturbation consistency across diverse scenes, we design a cross-view adversarial alignment mechanism to process multi-view images. Experiments demonstrate that MVAP-G significantly degrades VGGT performance without iterative optimization during inference. This work pioneers multi-view adversarial attacks on 3D foundation models, uncovering severe vulnerabilities and underscoring the urgent need for robust 3D vision systems. The code is available at \url{https://github.com/qsong2001/mvap-g}.

\keywords{Adversarial Examples \and VGGT \and 3D Safety}

\end{abstract}

%% file: sec/1_introduction.tex
\section{Introduction}
Recent advances in 3D vision have led to new 3D foundation models like Visual Geometry Grounded Transformer (VGGT)~\cite{wang2025vggt}. It unifies diverse 3D tasks like 3D reconstruction, SLAM, tracking~\cite{deng2025vggt-long,wang2025pi,shen2025fastvggt,wang2025fastervggt} into a single feed-forward pass. However, as these models are increasingly deployed in safety-critical applications~\cite{maggio2025vggt-slam, xiao2025spatialtrackerv2}, their vulnerability to invisible adversarial perturbations emerges as a fundamental security concern~\cite{song2026creating, luo2026adversarially}.

\begin{figure}[!t]
  \centering
  \includegraphics[width=1.00\linewidth]{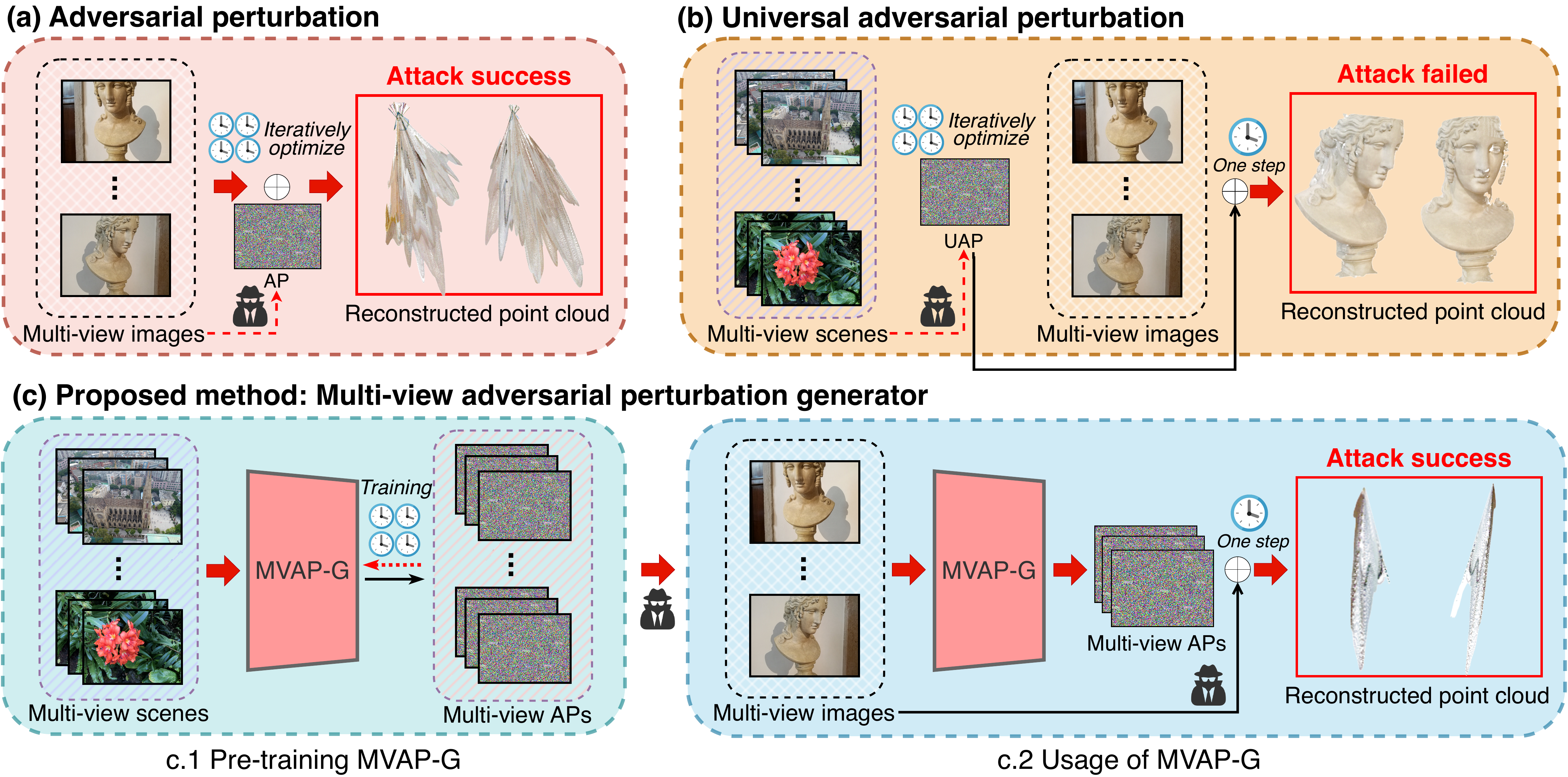}
  \caption{Different adversarial examples against VGGT. Although traditional (a) adversarial perturbation (AP)~\cite{madry2018towards} can achieve effective attack performance, it must be optimized for each scene with significant computational cost. (b) Universal adversarial perturbation~\cite{moosavi2017universal} (UAP) is optimized on multi-view scenes data, yet it lacks attack effectiveness. (c) Our proposed MVAP-G is trained on multi-view data and can generate multi-view perturbations (Multi-view APs) against VGGT in real-time. }
  \label{fig:teaser}
  \vspace{-0.20cm}
\end{figure}

Adversarial attacks are a major threat to the security of most vision applications~\cite{madry2018towards,xie2017adversarial,poursaeed2018generative,baluja2017adversarial,zhou2024darksam}. However, large foundation models like VGGT~\cite{wang2025vggt}, with their massive parameter scales, exhibit robustness to naive adversarial perturbations~\cite{madry2018towards} in \emph{real-world applications}. Traditional adversarial attacks~\cite{goodfellow2014explaining_fgsm,madry2018towards} require continuous per-scene optimization for every new multi-view input (as shown in Fig.~\ref{fig:teaser}). This computational overhead makes them impractical in dynamic environments where scenes change continuously in the real world. To eliminate per-scene optimization, Universal Adversarial Perturbations (UAPs)~\cite{moosavi2017universal,zhou2024darksam} employ data-agnostic, universal perturbations applicable to all inputs. This ``one-for-all'' strategy eliminates iterative computation, offering high efficiency. However, UAPs fail against VGGT due to their limited expressiveness: a static perturbation cannot adapt to the rich geometric variations in multi-view scenes~\cite{zhang2025improving,hu2024generate,huang2024texture}. As a result, UAPs exhibit poor capability against VGGT across diverse scenes.

A promising direction is to leverage generative models that can dynamically synthesize adversarial perturbations in real time. Prior generative adversarial example methods~\cite{hayes2018learning,poursaeed2018generative,baluja2017adversarial} have demonstrated success in this paradigm, efficiently creating perturbations through learned generators without requiring per-instance optimization. However, these methods are fundamentally designed for single-view architectures and cannot effectively handle multi-view inputs. They fail to model cross-view consistency and geometric relationships, which are critical factors for attacking multi-view foundation models like VGGT that explicitly exploit geometric correspondences across views.

To address this limitation, we propose \textbf{M}ulti-\textbf{V}iew \textbf{A}dversarial \textbf{P}erturbations \textbf{G}enerator (\textbf{MVAP-G}), the first generative approach specifically designed for attacking multi-view 3D foundation models. MVAP-G directly processes raw multi-view images and produces imperceptible yet geometrically consistent adversarial perturbations across all views in real time. By learning to generate view-coherent perturbations, MVAP-G combines the efficiency of generative methods with the geometric awareness required for multi-view scenarios, requiring no test-time optimization while adapting seamlessly to diverse scene configurations.

We achieve this through three core designs. 
\textbf{First}, a cross-view adversarial alignment (CAA) module explicitly models inter-view correspondences, forcing the generated perturbations to remain visually imperceptible while being destructive to VGGT's inference. 
\textbf{Second}, dynamic perturbation regularization is applied during training to stabilise optimization under strong attack objectives, balancing attack performance and the invisibility of generated perturbations. \textbf{Third}, we demonstrate that pre-training the generator on a single-view dataset~\cite{reizenstein2021common} can reduce total training costs. This simple yet effective strategy learns rich prior knowledge about adversarial vulnerability against VGGT. With these designs, MVAP-G consistently reduces VGGT's performance in real time and across unseen scenes. Our contributions can be summarized:

\begin{itemize}
    \item The first study on adversarial examples of the visual geometry grounded transformer, exploiting and revealing the potential vulnerabilities in the visual geometry grounded transformer.
    \item A real-time multi-view perturbation generation method that dynamically injects adversarial perturbations onto multi-view images, markedly surpassing prior adversarial perturbation methods.
    \item This work establishes a new paradigm for adversarial robustness in 3D vision, emphasizing the urgent need for secure deployment strategies of 3D foundation models.
\end{itemize}
Extensive experiments demonstrate the effectiveness and generalizability of our method across various input scenes. 

%% file: sec/2_related_work.tex
\section{Related work}
\subsection{Universal adversarial perturbations}
Adversarial examples challenge the security of deep learning systems. These carefully crafted perturbations cause models to misclassify without being perceptible to humans. Early methods such as the Fast Gradient Sign Method (FGSM)~\cite{goodfellow2014explaining_fgsm} and Projected Gradient Descent (PGD)~\cite{madry2018towards} use gradient-based optimization to generate instance-specific perturbations. They expose the vulnerability of neural networks~\cite{goodfellow2013maxout,he2016deep, xie2025towards} to manipulated inputs. However, these attacks require iterative optimization per input, which proves computationally expensive at scale. This limitation motivates the study of Universal Adversarial Perturbations (UAPs)~\cite{moosavi2017universal}. UAPs enable one-shot attacks via a single, precomputed perturbation. Unlike sample-specific attacks, a UAP deceives the model across diverse inputs, revealing systemic risks in deployment. The original UAP~\cite{moosavi2017universal} aggregates instance-specific directions iteratively and fools ImageNet-trained networks~\cite{deng2009imagenet}. Later work branches into two directions: generative methods use GANs to improve transferability across architectures~\cite{hayes2018learning,poursaeed2018generative,zhao2022learning}, while gradient-based approaches optimize in feature subspaces~\cite{zhou2024darksam,lin2020nesterov,wu2022small}. Recent studies highlight limitations of CNN-focused UAPs on vision transformers~\cite{lu2025uap, huang2025x_transfer, zhang2026semanticaware}. New paradigms emerge, including geometric transformations~\cite{kanbak2018geometric} and attacks on attention mechanisms~\cite{liu2024disrupting}. Nonetheless, conventional scene-agnostic UAPs falter in multi-view, geometry-aware tasks like 3D reconstruction. They lack sufficient capacity and explicit cross-view modeling.

\subsection{Adversarial attack in 3D}
3D foundation models transform multi-view reconstruction tasks. However, their complex architectures introduce unique adversarial vulnerabilities. Traditional $\ell_p$-bounded perturbations effectively disrupt 2D CNNs with imperceptible noise that alters classification outputs~\cite{huang2024texture}. Similarly, attacks like NeRFool~\cite{fu2023nerfool} target generalizable NeRFs~\cite{wang2021ibrnet,chen2021mvsnerf}. These rely on iterative optimization, which is incompatible with feed-forward 3D models due to pipeline differences.
Universal Adversarial Perturbations (UAPs)~\cite{moosavi2017universal} improve efficiency via input-agnostic noise. However, they fail to generalize across diverse multi-view inputs in 3D tasks. They lack mechanisms to handle geometric and temporal dependencies~\cite{lu2025uap}. For instance, NerFail~\cite{jiang2024nerfail} applies multi-view attacks on NeRFs but struggles with varying viewpoints. This highlights the limitations of static perturbations in dynamic 3D settings. More broadly, INR-based representations have also been studied for security-oriented information hiding and steganography, reflecting growing interest in the security of neural scene representations~\cite{song2026securing,song2025unified}.
Recent work explores more targeted 3D attacks. Poison-splat~\cite{lu2025poison} poisons input images to increase training costs in 3D Gaussian Splatting (3DGS)~\cite{kerbl20233d}. MPAM-3DGS~\cite{jiang2025mpam} manipulates Gaussian parameters to degrade reconstruction quality. Targeted attacks~\cite{horvath2023targeted} on generalizable NeRFs mislead downstream tasks like object detection~\cite{chen2018shapeshifter,xie2017adversarial,lu2017no, han2026quantum}. Recent studies have investigated copyright protection for 3D representations through reconstruction prevention and ownership verification~\cite{song2024geometry, song2024protecting, luo2023copyrnerf,han2026gausstrace,han2026gs_checker}. Despite their effectiveness, most methods require iterative optimization~\cite{madry2018towards}. This incurs a high computational cost with multi-view inputs and large models such as VGGT~\cite{wang2025vggt}. Consequently, they remain impractical for real-world deployment.

\begin{figure*}[!t]
  \centering
  \includegraphics[width=1.0\linewidth]{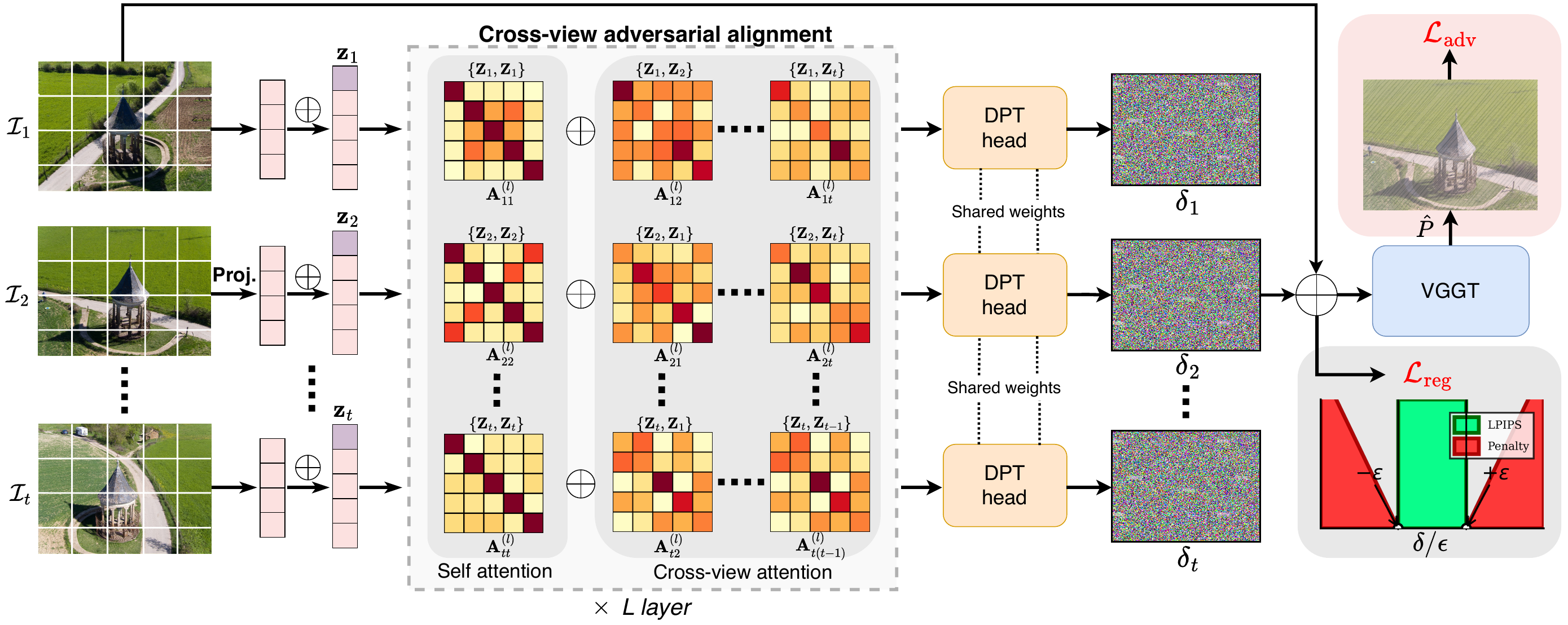}
  \vspace{-0.6cm}
 \caption{Overview of the Multi-view Adversarial Perturbation Generator (MVAP-G). The generator $\mathcal{G}$ takes a multi-view image sequence $\mathcal{S}$ and produces perturbations $\delta_t$ for each frame in a single feed-forward pass. The perturbed images $\hat{I}_t = I_t + \delta_t$ are then fed into the pre-trained VGGT model $\mathcal{F}$ to produce adversarial 3D point clouds $\hat{P}$. The generator is built on a transformer architecture with cross-view adversarial alignment (CAA) and a DPT head for generating cross-view pixel-level perturbations.}
  \label{fig:framework}
\end{figure*}

\section{Background}
\noindent \textbf{Visual geometry grounded transformer.}
VGGT~\cite{wang2025vggt} leverages a vision foundation transformer model~\cite{oquab2025dinov2} combined with DPT~\cite{ranftl2021vision} to achieve 3D scene reconstruction inference in seconds~\cite{deng2025vggt-long}. VGGT integrates tasks such as camera pose estimation, depth prediction, and point cloud reconstruction in a single feed-forward pass, using cross-view feature consistency to generate viewpoint-invariant 3D representations. Formally, VGGT can be regarded as a function \(\mathcal{F}\) that processes a multi-view image sequence \(\mathcal{S} = \{I_t \in \mathbb{R}^{H \times W \times 3}\}_{t=1}^T\) to produce geometric outputs as:
\begin{equation}
\mathbf{Y} = \mathcal{F}(\mathcal{S}; \phi),
\end{equation}
where \(\phi\) denotes the parameters of the pretrained vision transformer~\cite{caron2021emerging} and DPT backbone~\cite{ranftl2021vision}. \(\mathbf{Y} = (P, \{D_t\}_{t=1}^T, \{\boldsymbol{\theta}_t\}_{t=1}^T)\) contains the 3D attributes, including point clouds \(P \in \mathbb{R}^{N \times 3}\), depth maps \(\{D_t \in \mathbb{R}^{H \times W}\}_{t=1}^T\), and camera poses \(\{\boldsymbol{\theta}_t \in \mathbb{R}^6\}_{t=1}^T\). VGGT effectively processes multi-view images and produces high-quality 3D content in a single forward pass, establishing a new stage in robust 3D foundation modeling.

\noindent \textbf{Problem setup.}  
VGGT enables unified, feed-forward reconstruction of 3D attributes from arbitrary multi-view image sequences, powering downstream applications such as SLAM~\cite{maggio2025vggt-slam}, spatial tracking~\cite{xiao2025spatialtrackerv2}, and real-time 3D inference~\cite{deng2025vggt-long,shen2025fastvggt,wang2025fastervggt}.  
However, adversarial attacks pose a critical threat in real-world deployments, where subtle perturbations can distort depth maps or camera poses, undermining the safety of safety-critical systems.  

Formally, given a clean multi-view sequence \(\mathcal{S} = \{I_t \in \mathbb{R}^{H \times W \times 3}\}_{t=1}^T\), an adversarial perturbation \(\boldsymbol{\delta}\) is added to induce significant errors in VGGT's output \(\mathcal{F}(\mathcal{S})\).  
The attack objective is defined as:
\begin{equation}
\begin{aligned}
\boldsymbol{\delta}^* = \arg \max_{\boldsymbol{\delta}} \ \mathcal{L}_{\text{adv}}\big(\mathcal{F}(\mathcal{S} + \boldsymbol{\delta}), \mathcal{F}(\mathcal{S})\big), 
\quad \text{s.t.} \quad \|\boldsymbol{\delta}\|_p \leq \epsilon,
\end{aligned}
\end{equation}
where \(\epsilon\) enforces imperceptibility under an \(\ell_p\)-norm constraint, and \(\mathcal{L}_{\text{adv}}\) quantifies the deviation in 3D outputs.

Traditional methods like FGSM~\cite{goodfellow2014explaining_fgsm} and PGD~\cite{madry2018towards} require per-scene iterative optimization, rendering them impractical for real-time, dynamic environments.  
Universal Adversarial Perturbations (UAPs)~\cite{moosavi2017universal} eliminate iteration by applying a fixed pattern across inputs, but fail to model multi-view geometry, resulting in poor transferability against transformer-based 3D models like VGGT.

These shortcomings highlight the need for single-pass perturbations that preserve cross-view consistency and scale to real-world conditions. Our MVAP-G framework meets this demand by generating input-conditioned, aligned cross-view adversarial noise in a single feed-forward pass. This design enables efficient, transferable attacks on VGGT in real-time deployment scenarios.

%% file: sec/3_method.tex
\section{Proposed method}
The key challenge in generating adversarial examples for VGGT~\cite{wang2025vggt} is creating perturbations that are both effective and computationally efficient. We address this with the Multi-view Adversarial Perturbation Generator (MVAP-G), which learns an input-conditioned mapping to generate multi-view adversarial examples against VGGT in a single feed-forward pass (as shown in Fig.~\ref{fig:framework}). 

\subsection{Multi-view perturbation generator}
MVAP-G takes a multi-view image sequence \(\mathcal{S} = \{I_t \in \mathbb{R}^{H \times W \times 3}\}_{t=1}^T\) as input and generates adversarial examples in a single feed-forward pass. These perturbations are optimized to disrupt the 3D attribute point clouds \(P \in \mathbb{R}^{N \times 3}\) generated by the pretrained VGGT model \(\mathcal{F}\). The adversarial objective is formulated to maximize the loss function \(\mathcal{L}_{\text{adv}}\), thereby degrading the performance of VGGT $\mathcal{F}$:
\begin{equation}
\boldsymbol{\delta}^* = \arg \max_{\{\delta_t\} \in \Delta} \mathcal{L}_{\text{adv}}\left(\mathcal{F}(\{I_t + \delta_t\}), \mathcal{F}(\mathcal{S})\right),
\end{equation}
\noindent where the perturbations are generated via the MVAP-G generator \(\mathcal{G}\) as:
\begin{equation}
\{\delta_t\} = \mathcal{G}(\mathcal{S}; \mathbf{w}),
\end{equation}
\noindent with $\mathbf{w}$ representing the learnable parameters of the generator. The constraint set \(\Delta\) ensures imperceptibility by enforcing \(\|\delta_t\|_\infty \leq \epsilon\).

The MVAP-G is built on a transformer architecture~\cite{dosovitskiy2021image,oquab2025dinov2,ranftl2021vision} with introduced cross-view adversarial alignment (CAA) and a DPT head for generating cross-view pixel-level perturbations. For an input sequence \(\mathcal{S}\), each frame \(I_t\) is divided into \(N_p\) \(16\times16\) patches, projected into latent vectors \(\mathbf{e} \in \mathbb{R}^{D}\) (\(D=384\)). Spatio-temporal relationships are encoded via \textit{learnable} camera embeddings \(\mathbf{p}_t\), yielding:
\begin{equation}
    \mathbf{z}^{(0)}_t = [ \mathbf{e}_{t,1}, \mathbf{e}_{t,2}, \ldots, \mathbf{e}_{t,N_p} ] + \mathbf{p}_t,
\end{equation}
where \(\mathbf{e}_{t,i}\) denotes the embedding vector of the \(i\)-th patch in the \(t\)-th view, and \([\cdot]\) represents the concatenation operation. These initial embeddings, $\mathbf{z}^{(0)}_t$, are then processed through multiple transformer layers, each incorporating the proposed cross-view adversarial alignment mechanism to enable information exchange across views.

\noindent \textbf{Cross-view adversarial alignment.}  
A cross-view adversarial alignment (CAA) module is proposed to facilitate understanding across multiple views. The attention mechanism combines both self-attention within each view and cross-attention across different views:
\begin{equation}
    \mathbf{z}^{(l)}_t = \text{LayerNorm}( \mathbf{z}^{(l-1)}_t + \mathbf{A}^{(l)}_t ),
\end{equation}
\noindent where the attention output $\mathbf{A}^{(l)}_t$ for view $t$ is the sum of attention contributions from all views:
\begin{equation}
\mathbf{A}^{(l)}_t = \sum_{k=1}^T \mathbf{A}^{(l)}_{tk},
\end{equation}
\noindent with each component computed as:
\begin{equation}
\mathbf{A}^{(l)}_{tk} = \text{softmax}(\frac{\mathbf{z}^{(l-1)}_t (\mathbf{z}^{(l-1)}_k)^{\top}}{\sqrt{D}})\mathbf{z}^{(l-1)}_k.
\end{equation}
When $t = k$, $\mathbf{A}^{(l)}_{tt}$ represents self-attention within the same view. When $t \neq k$, $\mathbf{A}^{(l)}_{tk}$ represents cross-view attention between different views. This formulation enables comprehensive information exchange across the multi-view sequence. After processing the multi-view features through the transformer layers with CAA, the resulting feature representations are passed to the Dense Prediction Transformer (DPT) head to generate pixel-level perturbations.

\noindent \textbf{Dense prediction transformer head.}
To obtain pixel-level perturbations from patch tokens, we adopt a Dense Prediction Transformer (DPT) head~\cite{ranftl2021vision}. Specifically, we extract features from the 3rd, 6th, 9th, and 12th transformer layers using four learned readout projections:
\begin{equation}
\mathbf{h}^{(l)}_t = \mathbf{z}_t^{(l)} W_{\text{readout}}^{(l)} 
\in \mathbb{R}^{N_p \times D}, 
\quad l \in \{3,6,9,12\},
\end{equation}
where $N_p = (H/16) \times (W/16)$ is the number of patches, and $D{=}384$ is the embedding dimension. Following DPT~\cite{ranftl2021vision,yang2024depth}, these multi-scale features are progressively fused through four Reassemble stages ($s{=}1{\sim}4$). Each stage upsamples to resolution $H/2^s \times W/2^s$ by concatenating the input feature with $K_s \in \{1,4,16,64\}$ learnable fusion tokens and applying a linear projection.   

After the final stage, a lightweight refinement module (3×3 depth-wise + 1×1 convolution) outputs the unbounded perturbation:
\begin{equation}
\tilde{\delta}_t = \text{Refine}\bigl(\mathbf{r}^{(4)}_t\bigr) 
\in \mathbb{R}^{H \times W \times 3},
\end{equation}
where $\mathbf{r}^{(4)}_t \in \mathbb{R}^{H \times W \times D}$ denotes the full-resolution feature map. The perturbation is constrained \textit{only during the inference} within the $\ell_\infty$-ball via:
\begin{equation}
\delta_t = \text{clip}(\tilde{\delta}_t, -\epsilon, \epsilon),
\label{eq:clip}
\end{equation}
which guarantees $\|\delta_t\|_\infty \leq \epsilon$ by clipping the unbounded perturbation to the predefined budget. MVAP-G adopts a multi-scale feature-decoding~\cite{ranftl2021vision} strategy similar to VGGT to process hierarchical features. This enables stable pixel-level perturbation generation, producing visually imperceptible noise that remains perfectly consistent across all views.

%%%%%%%%%%%%%%%%%%%%%%%%%%%%%%%%%%%%%%%%

\input{fig/algorithm}

%%%%%%%%%%%%%%%%%%%%%%%%%%%%%%%%%%%%%%%%

\subsection{Training objective}

Given a pre-trained VGGT model $\mathcal{F}$ and our MVAP-G generator $\mathcal{G}$, the training objective follows a step-by-step adversarial process to optimize the generator $\mathcal{G}$ for disrupting the 3D reconstruction model \(\mathcal{F}\). First, each input image \(I_t\) from the multi-view sequence \(\mathcal{S} = \{I_t \in \mathbb{R}^{H \times W \times 3}\}_{t=1}^T\) is fed into the generator $\mathcal{G}$, parameterized by weights \(\mathbf{w}\), to produce per-frame perturbations $\{\delta_t\}_{t=1}^T$ via $\{\delta_t\} = \mathcal{G}(\mathcal{S}; \mathbf{w})$. The perturbed sequence $\hat{\mathcal{S}} = \{I_t + \delta_t\}_{t=1}^T$ is processed by VGGT to generate the adversarial output:
\begin{equation}
\hat{\mathbf{Y}} = \mathcal{F}(\hat{\mathcal{S}}).
\end{equation}
\noindent where $\hat{\mathbf{Y}} = (\hat{P}, \{\hat{D}_t\}_{t=1}^T, \{\hat{\boldsymbol{\theta}}_t\}_{t=1}^T)$. 
We target the output point cloud $\hat{P}$ as the primary attack objective, given its central role in VGGT's 3D reconstruction pipeline~\cite{wang2025vggt} (We also discuss the impacts on other outputs of VGGT in \cref{sec:other_impacts}). To effectively degrade the output, the adversarial loss \(\mathcal{L}_{\text{adv}}\) is formulated as the mean L2 norm of the adversarial point cloud coordinates, designed to shrink the point cloud to the origin $(0,0,0)$~\cite {song2024geometry}:
\begin{equation}
\mathcal{L}_{\text{adv}} = \mathbb{E}_{\mathcal{S} \sim \mathcal{D}_{\text{train}}} [ \frac{1}{|\hat{P}|} \sum_{p \in \hat{P}} \|p\|_2 ].
\label{eq:adv_loss}
\end{equation}
This simple yet effective metric efficiently disrupts the 3D structure (further discussed in \cref{sec:adv_loss}).

To constrain the generated perturbations $\delta$, we apply a multi-component regularization loss \textit{during the training}:
\begin{equation}
\mathcal{L}_{\text{reg}} = \sum_{t=1}^{T} \left[ \text{LPIPS}(\hat{I}_t, I_t) + \operatorname{ReLU}( \|\delta_t\|_\infty - \epsilon ) \right],
\label{eq:reg_loss}
\end{equation}
\noindent where $\text{LPIPS}(\cdot,\cdot)$~\cite{zhang2018unreasonable} measures perceptual similarity in deep feature space. The ReLU term enforces the hard $\ell_\infty$ constraint $\|\delta_t\|_\infty \leq \epsilon$. 

The total objective combines both terms with a balancing coefficient $\lambda$:
\begin{equation}
\mathcal{L}_{\text{total}} = \mathcal{L}_{\text{adv}} + \lambda \mathcal{L}_{\text{reg}}.
\end{equation}
The generator $\mathcal{G}$ parameters $\mathbf{w}$ are optimized using AdamW~\cite{loshchilov2019adaw} to minimize $\mathcal{L}_{\text{total}}$ over the training dataset $\mathcal{D}_{\text{train}}$, which contains diverse multi-view sequences for robust perturbation learning.

% The training dataset $\mathcal{D}_{\text{train}}$ comprises diverse multi-view sequences, enabling $\mathcal{G}$ to learn robust perturbations that generalize across varied scenes. By optimizing the adversarial objective over this dataset, MVAP-G learns to generate effective perturbations in a single feed-forward pass that significantly degrade 3D reconstruction quality while maintaining imperceptibility through careful balancing of $\mathcal{L}_{\text{adv}}$ and $\mathcal{L}_{\text{reg}}$. 

\subsection{Implementation details}
MVAP-G undergoes a two-stage training procedure. First, the model is pre-trained on single-view images from the COCO dataset~\cite{lin2014microsoft} for 4 million iterations using AdamW~\cite{loshchilov2019adaw} optimizer with a learning rate of $1\times10^{-5}$. The regularization loss is introduced after 2.4 million iterations to allow stable feature learning. This single-view pre-training enables the generator to learn robust image features for effective perturbation generation before multi-view adaptation. Subsequently, the model is fine-tuned on multi-view imagery. Each training batch samples 2-10 frames randomly from a scene. The multi-view training corpus comprises seven diverse 3D datasets: CO3Dv2~\cite{reizenstein2021common}, BlendMVS~\cite{yao2020blendedmvs}, ScanNet~\cite{dai2017scannet}, Virtual KITTI~\cite{gaidon2016virtual}, DTU~\cite{jensen2014large}, ETH3D~\cite{schops2017multi}, and FlyingThings3D~\cite{mayer2016large}. These datasets encompass both indoor/outdoor environments and synthetic/real-world data, providing diversity similar to MASt3R~\cite{leroy2024grounding} and VGGT~\cite{wang2025vggt}. Following VGGT's preprocessing procedure, input frames are resized to a maximum dimension of 518 pixels. The complete training requires approximately 20 days on a single NVIDIA RTX 4090 GPU. More implementation details can be found in the \textit{supplemental material}.

%% file: fig/algorithm.tex
\begin{algorithm}[t]
   \caption{MVAP-G training procedure}
   \label{alg:mvapg}
   \SetKwInOut{Input}{Input}
   \SetKwInOut{Output}{Output}
   
   \Input{VGGT model $\mathcal{F}$, multi-view image dataset $\mathcal{D}_{\text{train}}$, Learning rate $\alpha$, Perturbation bound $\epsilon$, regularization coefficient $\lambda$, number of iterations $N_{\text{iter}}$}
   \Output{MVAP-G ($\mathcal{G}$) parameters $\mathbf{w}$}
   
 Initialize MVAP-G generator: $\mathcal{G}_{\mathbf{w}}$\;

   \For{$\text{iter}$ in $0$ to $N_{\text{iter}}$}{
      \tcp{Batch multi-view sequences}
      $\mathcal{S}_{\text{batched}} \gets \text{Batch}(\mathcal{D}_{\text{train}})$\;
      
     \tcp{Generate perturbations}
     $\Delta = \{\delta_t\}_{t=1}^T \gets \mathcal{G}_{\mathbf{w}}(\mathcal{S})$\;
     
     \tcp{Create perturbed sequence}
     $\hat{\mathcal{S}} \gets \{I_t + \delta_t\}_{t=1}^T$\;
     
     \tcp{Compute adv. point cloud}
     $\hat{P} \gets \mathcal{F}(\hat{\mathcal{S}})$\;
     
     \tcp{Compute adv. loss}
     $\mathcal{L}_{\text{adv}} \gets \frac{1}{|\hat{P}|} \sum_{p \in \hat{P}} \|p\|_2$\;
     
     \tcp{Compute regularization loss}
     $\mathcal{L}_{\text{reg}} \gets \text{ReLU}(\cdot) + \text{LPIPS}(\cdot)$\;

     \tcp{Compute total loss}
     $\mathcal{L}_{\text{total}} \gets \mathcal{L}_{\text{adv}} + \lambda \mathcal{L}_{\text{reg}}$\;

     \tcp{Update parameters}
     $\mathbf{w} \gets \mathbf{w} - \alpha \cdot \nabla_{\mathbf{w}} (\mathcal{L}_{\text{total}})$\;
   }
   \Return $\mathbf{w}$\;
\end{algorithm}

%% file: sec/4_experiments.tex
\section{Experiments}
\noindent \textbf{Setup.} To evaluate the effectiveness of our proposed MVAP-G framework, we conduct extensive experiments targeting the Visual Geometry Grounded Transformer (VGGT)~\cite{wang2025vggt}. We assess the attack performance on point cloud reconstruction and related tasks, measuring degradation using the Chamfer Distance (CD)~\cite{barrow1977parametric} for point clouds and the Mean Squared Error (MSE) for the reconstructed point cloud. The evaluation spans four datasets: COCO~\cite{lin2014microsoft}, a large-scale dataset with single-view RGB images to test generalization to real-world scenes; ImageNet~\cite{deng2009imagenet}, a diverse single-view dataset to evaluate perturbation transferability across object categories; Co3D~\cite{reizenstein2021common}, a real-world multi-view dataset contains common 3D object; and LLFF~\cite{mildenhall2019llff}, a real-world multi-view dataset capturing dynamic scenes with complex lighting and geometry.

\noindent \textbf{Benchmarks.} MVAP-G is compared against several baseline methods: Random Noise, which adds Gaussian noise to input images as a naive baseline; Universal Adversarial Perturbations (UAP)~\cite{moosavi2017universal}, a scene-agnostic attack method; and naive adversarial attacks, \ie, AP ($\text{iter}{=}10$ and $\text{iter}{=}20$)~\cite{madry2018towards}, optimized with steps of 10 and 20 iterations, respectively. We evaluate performance using Chamfer Distance (CD)~\cite{barrow1977parametric}. We report the shifted CD distance between outputs from perturbed and clear multi-view images. For multi-view datasets, performance is assessed across varying numbers of views $v \in \{1,4,8,16,25\}$. Computational efficiency is measured by the inference time and peak GPU memory per view.

\input{fig/exp-table-main}
\input{fig/exp-fig-main}

\subsection{Main results}

\noindent \textbf{Attack performance.} Table~\ref{tab:main} compares MVAP-G against baselines on point cloud reconstruction. Random noise yields a negligible Chamfer Distance (CD), confirming its ineffectiveness. UAP~\cite{moosavi2017universal} improves over noise but degrades sharply in multi-view settings due to the absence of cross-view modeling. Naive adversarial perturbation, AP ($\text{iter}{=}10$, $\text{iter}{=}20$)~\cite{madry2018towards} achieves stronger disruption, with marginal gains from additional iterations. In contrast, MVAP-G consistently outperforms all baselines across datasets and view counts, achieving the highest CD in every configuration using only a single feed-forward pass. Notably, on multi-view CO3D and LLFF (Fig.~\ref{fig:adv_sample}), MVAP-G maintains robust attack strength, far surpassing UAP and closely competing with or exceeding iterative AP, demonstrating effective exploitation of VGGT vulnerabilities. Additional results are available in the \textit{supplemental material}.

\noindent \textbf{Efficiency.} MVAP-G generates universal multi-view adversarial perturbations in a single feed-forward pass, achieving real-time inference and low memory footprint across increasing view counts. As shown in Fig.~\ref{fig:efficiency} (a), inference time scales sub-linearly with the number of views, remaining in the millisecond range even for large inputs.  In contrast, Naive AP~\cite{madry2018towards} requires per-scene iterative optimization~\cite {madry2018towards}, leading to exponential runtime growth and reaching minutes per scene at high view counts. Similarly, Fig.~\ref{fig:efficiency} (c) reveals that MVAP-G maintains significantly lower peak GPU memory than Naive AP, which exceeds 30 GB under heavy multi-view loads. These results highlight MVAP-G's superior scalability, enabling efficient, real-time attacks on VGGT without the computational overhead of scene-specific AP~\cite{madry2018towards}.

%%%%%%%%%%%%%%%%%%%%%%%%%%%%%%%%%%%%%%%%%%%%%%%%
%%%%% Ablation Study %%%%%
%%%%%%%%%%%%%%%%%%%%%%%%%%%%%%%%%%%%%%%%%%%%%%%%

\input{fig/exp-fig-effency}
\input{fig/exp-fig-abl}

\input{fig/exp-table-abl}

\input{fig/exp-adv-sample}

\subsection{Ablation studies}
\noindent \textbf{Component ablation.}  
To understand the contributions of MVAP-G's components, we conduct ablation studies to systematically evaluate the impact of each key module. Table~\ref{tab:component} reports incremental ablation results averaged over 1, 2, and 4 views. The baseline configuration uses a Conv+Linear decoder without pre-training, CAA, or $\mathcal{L}_\text{reg}$. Pre-training on COCO~\cite{lin2014microsoft} provides a boost in attack strength, as it can learn adversarial knowledge against VGGT. The DPT head outperforms the naive decoder by capturing richer features for pixel-level adversarial perturbations~\cite{ranftl2021vision}. CAA plays a limited role in increasing performance under limited review. However, it could enable robust performance in multi-view scenarios, with its benefits growing significantly as the number of views increases (as shown in Fig.~\ref{fig:abl} (a)). Finally, $\mathcal{L}_\text{reg}$ effectively minimizes perturbation size and ensures invisibility without compromising attack efficacy.

\noindent \textbf{Cross-view adversarial alignment (CAA).} 
Fig.~\ref{fig:abl} (a) shows attack performance (CD) versus input view count. Without CAA, performance degrades as view counts increase due to misalignment perturbations across views. In contrast, our full model with CAA leverages additional views to steadily enhance attack strength by exploiting multi-view geometry. These results demonstrate that CAA achieves competitive performance across multi-view scenarios. To further illustrate the effectiveness of CAA, we visualize results at view number $v{=}32$. \textit{Without CAA}, perturbations allow VGGT to partially recover geometry through multi-view attention fusion, yielding fragmented but recognizable structures.

\begin{figure}[!t]
    \centering
    \begin{minipage}[c]{0.65\linewidth}
        \centering
        \includegraphics[width=1.0\linewidth]{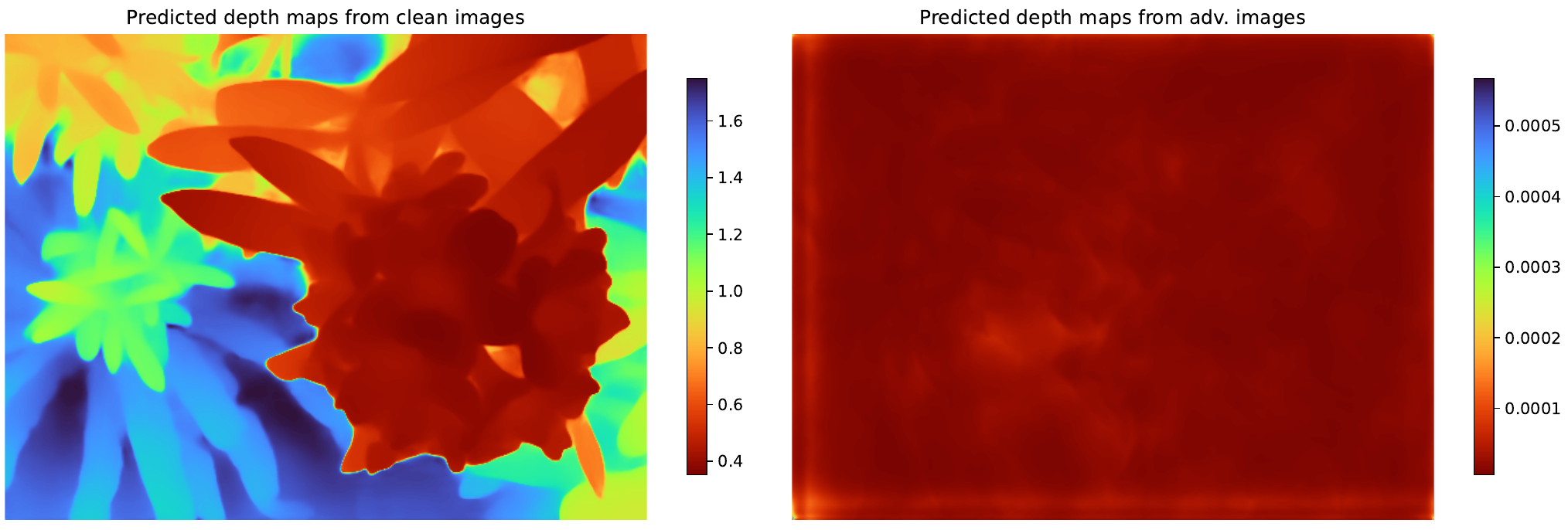}
    \end{minipage}
    \hfill
    \begin{minipage}[c]{0.30\linewidth}
        \centering
        \small
    \resizebox{1.0\linewidth}{!}{
        \begin{tabular}{lcc}
        \toprule
        Metric & Clean & Under Attack \\
        \midrule
        Rotation error ($^\circ$) & 2.3 & \textbf{44.4} \\
        Translation error (norm) & 0.05 & \textbf{0.43} \\
        Focal length shift (px) & -- & \textbf{18.7} \\
        \bottomrule
        \end{tabular}}
    \end{minipage}
    \vspace{-0.20cm}
    \caption{(left) Impacts on depth. (right) Impacts on camera pose. Camera pose under attack (CO3D $v\!=\!8$, avg. 100 scenes) (\textbf{\toblue{please zoom in for best review}}).}
    \label{fig:depth_pose}
    \vspace{-0.30cm}
\end{figure}

\noindent \textit{\textbf{Impact on Depth/Pose.}}
\label{sec:other_impacts}
Using the \textbf{same MVAP-G model} (trained \textit{only} on point cloud loss $\mathcal{L}_{\text{adv}}$), we observe significant degradation on VGGT's other geometric outputs (\ie, depth maps and camera poses) as shown in \cref{fig:depth_pose}. This occurs because VGGT's outputs share a common backbone that captures key geometric information.

\noindent \textbf{Regularization loss $\mathcal{L}_\text{reg}$.} To ensure the invisibility of the generated adversarial perturbation, the regularization loss $\mathcal{L}_\text{reg}$ was introduced after $2.4 \times 10^6$ training steps. Fig.~\ref{fig:abl} (b) compares training dynamics with and without $\mathcal{L}_\text{reg}$. Regularized training yields stable optimization, reliably reaching low $\mathcal{L}_{\text{adv}}$ with improving perturbation invisibility. By limiting perturbation magnitude, $\mathcal{L}_\text{reg}$ encourages precise, minimal noise that retains full attack strength, demonstrating that effective disruption of VGGT stems from targeted precision, not large-scale distortion. Fig.~\ref{fig:abl} (c) and visualized perturbation in Fig.~\ref{fig:adv_sample} further show that $\mathcal{L}_\text{reg}$ produces visually imperceptible, texture-preserving perturbations in input images, while effectively compromising VGGT's inference. More examples and analysis are provided in the \textit{supplemental material}.

\begin{figure}[t]
    \centering
    \centering
    \includegraphics[width=0.43\linewidth]{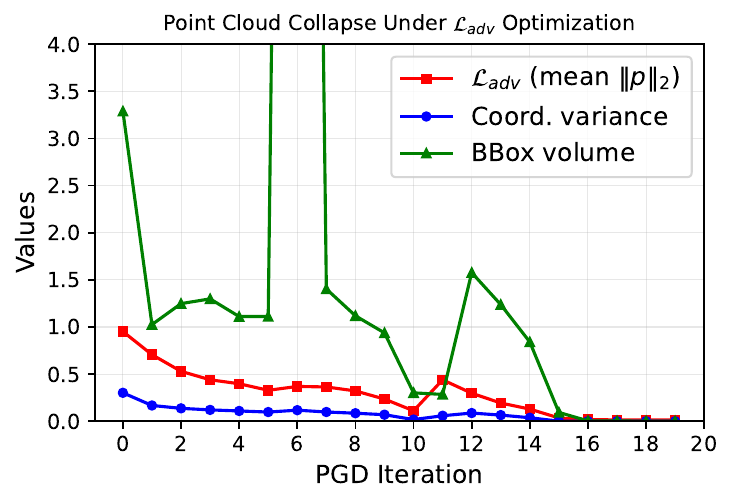}
    \includegraphics[width=0.43\linewidth]{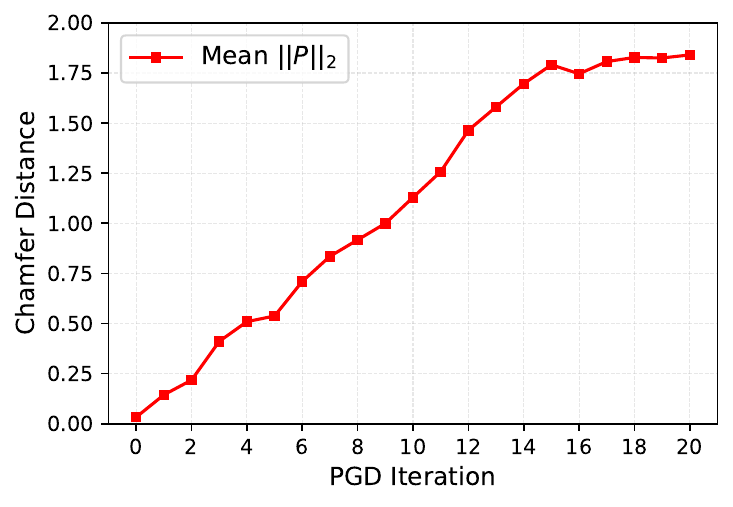}
    \vspace{-0.45cm}
    \caption{Statistics under $\mathcal{L}_\text{adv}$. (Left) The point cloud collapses to the origin $(0,0,0)$ via supervision of adversarial loss $\mathcal{L}_\text{adv}$. (Right) The CD distance between the reconstructed point cloud and the original increases in steps. }
    \vspace{-0.50cm}
    \label{fig:adv_loss}
\end{figure}

\noindent \textbf{Effectiveness of adversarial loss $\mathcal{L}_\text{adv}$.}
\label{sec:adv_loss}
We conduct a case study on a PGD optimization of  $\mathcal{L}_\text{adv}$ (\ie, min mean $||P||_2$). As shown in Fig. \ref{fig:adv_loss}, PGD optimization of $\mathcal{L}_\text{adv}$ increases the shifted CD from 0.02 to 1.82 within 20 iterations, while significantly reducing point cloud variance. This confirms that $\mathcal{L}_\text{adv}$ can effectively collapse the point cloud.

%% file: fig/exp-table-main.tex
\begin{table*}[!t]
\caption{Attack impacts on VGGT via different perturbation methods. We report Shifted CD distances between the reconstructed 3D point clouds from the clear and perturbed samples. Higher CD values denote a greater shifted distance and indicate better attack performance. MVAP-G achieves superior attack performance with \textbf{only a single feed-forward iteration}.}
\label{tab:main}
\centering
\resizebox{1.0\textwidth}{!}{
\begin{tabular}{lccccccccccccc}
\toprule
\multirow{2.5}{*}{Method} & 
COCO~\cite{lin2014microsoft} & ImageNet~\cite{deng2009imagenet} & 
\multicolumn{5}{c}{LLFF~\cite{mildenhall2019llff}} & 
\multicolumn{5}{c}{CO3D~\cite{reizenstein2021common}} \\
\cmidrule(lr){2-2}  \cmidrule(lr){3-3}  \cmidrule(lr){4-8} \cmidrule(lr){9-13}
 & \multicolumn{1}{c}{$v=1$} & \multicolumn{1}{c}{$v=1$} & 
 $v=1$ & $v=4$ & $v=8$ & $v=16$ & $v=25$  & 
 $v=1$ & $v=4$ & $v=8$ & $v=16$  & $v=25$  \\
\midrule
Gaus. noise    & 0.534 & 0.505 &0.912 & 0.900 & 0.941 & 0.949 & 0.982 & 0.678 & 0.666 & 0.671 & 0.632 & 0.672 \\
UAP            & 0.565 & 0.556 &0.940 & 0.964 & 0.980 & 1.010 & 1.043 & 0.673 & 0.675 & 0.671 & 0.673 & 0.659 \\
AP ($\text{iter}{=}10$)  & 1.361 & 1.349 &1.841 & 1.769 & 1.803 & 1.870 & 1.834 & 1.442 & 1.718 & 1.890 & 1.643 & 1.913\\
AP ($\text{iter}{=}20$)  & 1.535 & 1.584 &1.842 & 1.811 & 1.817 & 1.882 & \textbf{1.925} & 1.889 & 1.892 & 1.958 & 1.875 & 2.00 \\
\midrule 
\rowcolor{gray!20} MVAP-G ($\text{iter}{=}1$)       & \textbf{1.566} & \textbf{1.611} &\textbf{1.975}  &\textbf{1.885} &\textbf{1.857} &\textbf{1.922} &1.910 &\textbf{1.995} &\textbf{2.016} &\textbf{2.011} &\textbf{1.999} &\textbf{2.047} \\
\bottomrule
\end{tabular}
}
\end{table*}

%% file: fig/exp-fig-main.tex
\begin{figure*}[!t]
  \centering
  \includegraphics[width=1.0\linewidth]{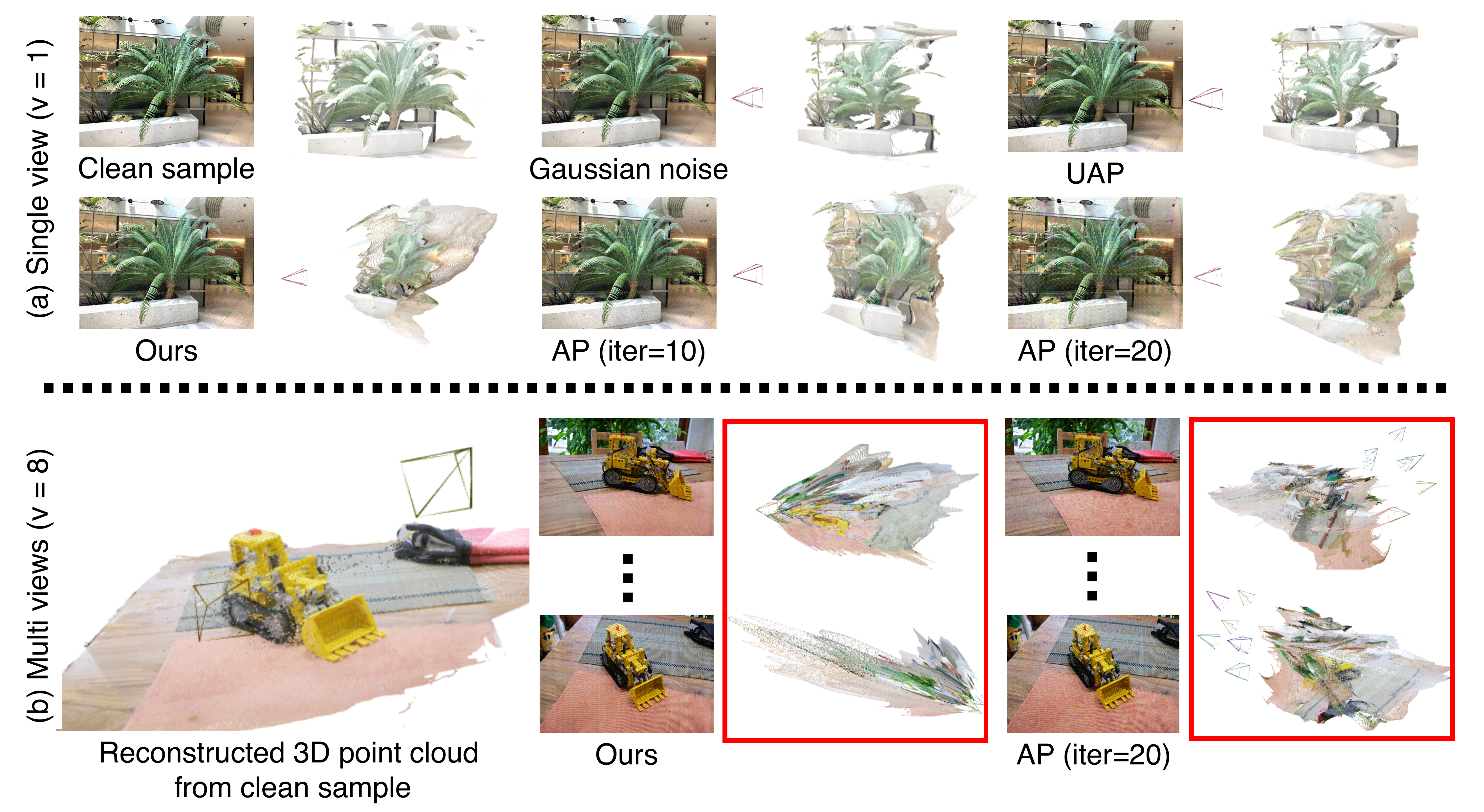}
      \vspace{-0.20cm}
  \caption{3D point cloud reconstruction under adversarial attacks for VGGT. (a) \textbf{Single-view ($v{=}1$)}: clean input and its ground truth; reconstructions from five adversarial perturbations, including Gaussian, UAP~\cite{moosavi2017universal}, AP with $\text{iter}{=}10$ and $\text{iter}{=}20$~\cite{madry2018towards}, and our MVAP-G. (b) \textbf{Multi-views ($v{=}8$)}: comparison between reconstructions from AP ($\text{iter}{=}20$) and our MVAP-G, demonstrating our method's superior effectiveness in a practical setting. Our MVAP-G causes severe degradation with a single pass. More results can be found in the supplemental material.}
  \label{fig:exp1}
\end{figure*}

%% file: fig/exp-fig-effency.tex
\begin{figure*}[!t]
  \centering
  \includegraphics[width=0.32\linewidth]
  {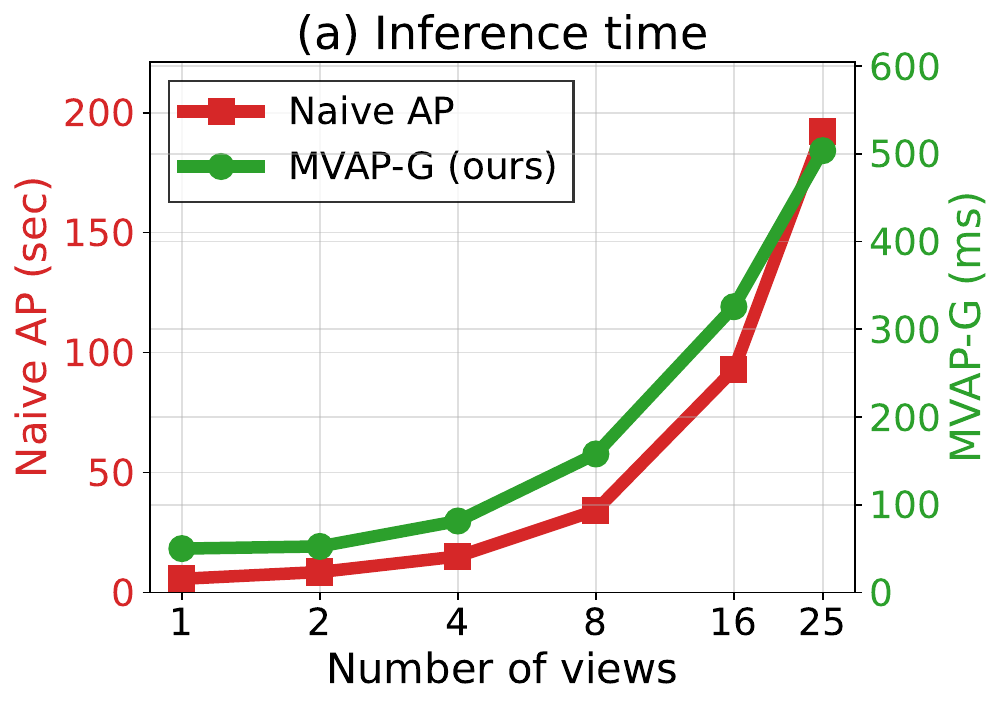}
  \hfill
  \includegraphics[width=0.32\linewidth]{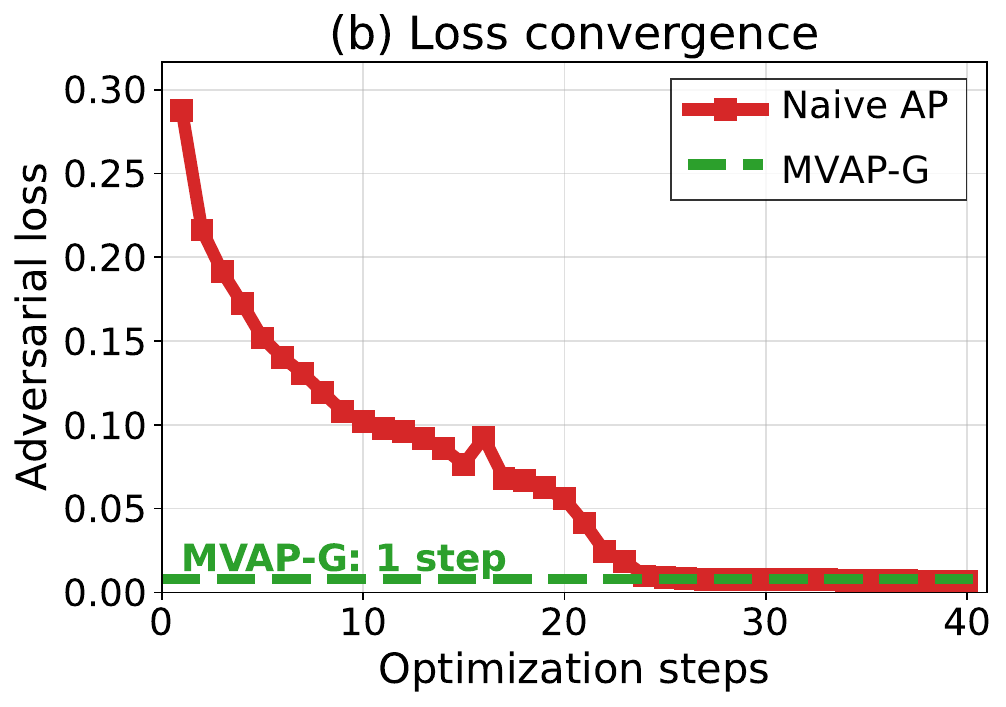}
    \hfill
\includegraphics[width=0.32\linewidth]{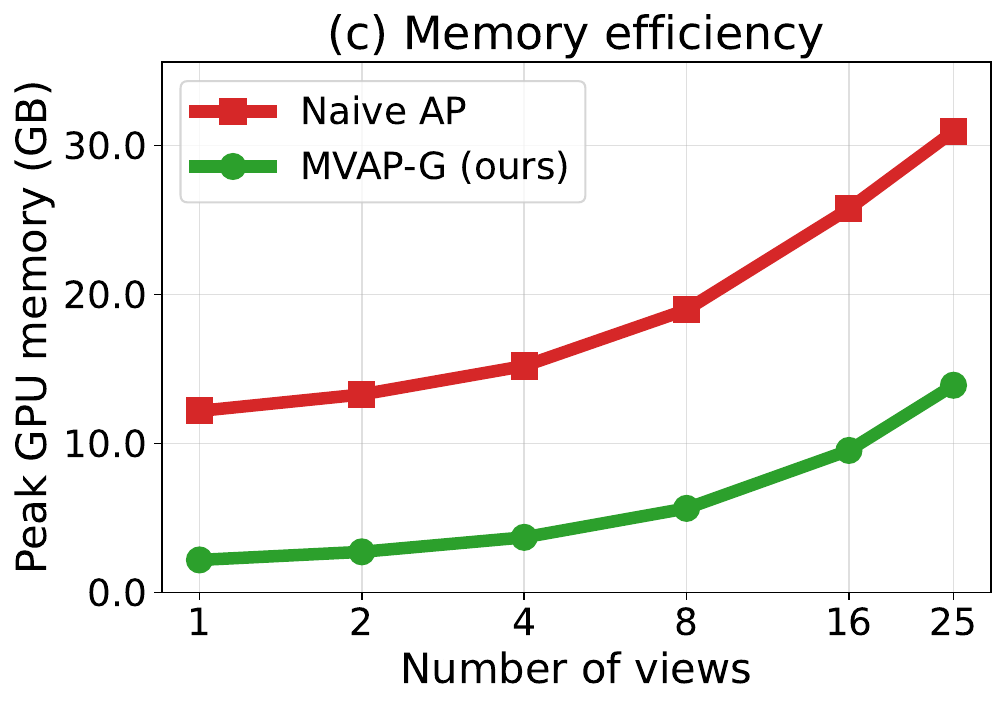}
  \caption{
      MVAP-G achieves real-time, memory-efficient, and high-quality multi-view attacks with a single feed-forward pass.
     \textbf{(a) Inference time} demonstrates that Naive AP requires over \textit{\magenta{25--200s}} per scene, while MVAP-G achieves millisecond-level attack in \textit{\magenta{50--500ms}} across multi-views input in a single forward pass. 
    \textbf{(b) Loss convergence} shows that Naive AP needs more than 25 steps to reach a loss of 0.008, whereas MVAP-G matches this performance in one step. 
    \textbf{(c) Memory efficiency} demonstrates that MVAP-G uses less than 14 GB of GPU memory, compared to ~30 GB for Naive AP. }
  \label{fig:efficiency}
\end{figure*}

%% file: fig/exp-fig-abl.tex
\begin{figure*}[!t]
  \centering
    \includegraphics[width=0.32\linewidth]{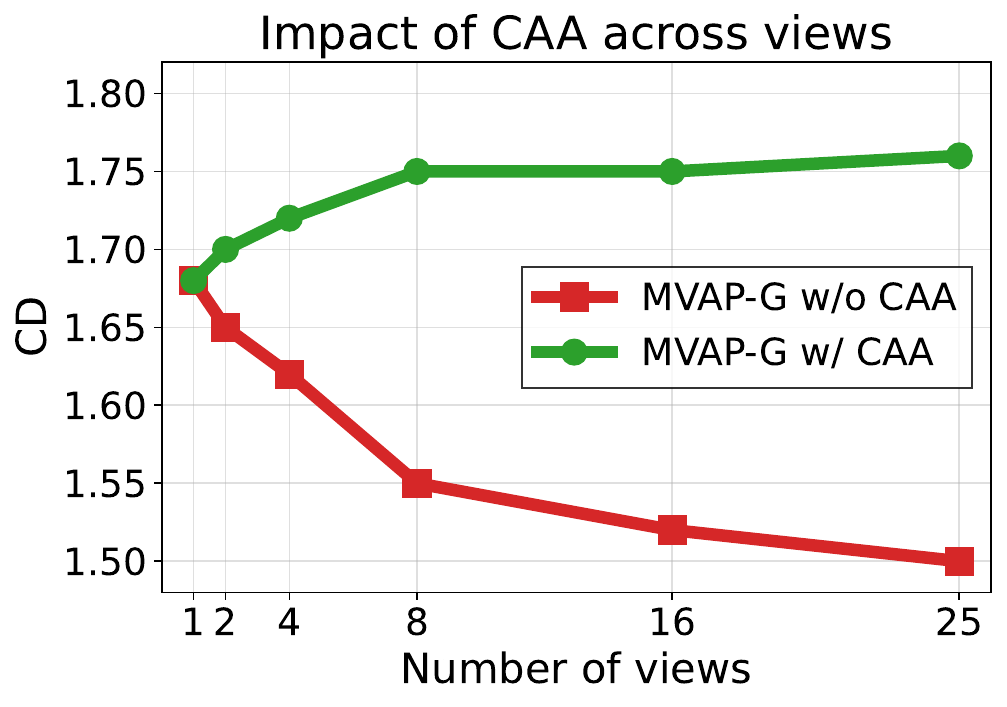}  
    \hfill
    \includegraphics[width=0.32\linewidth]{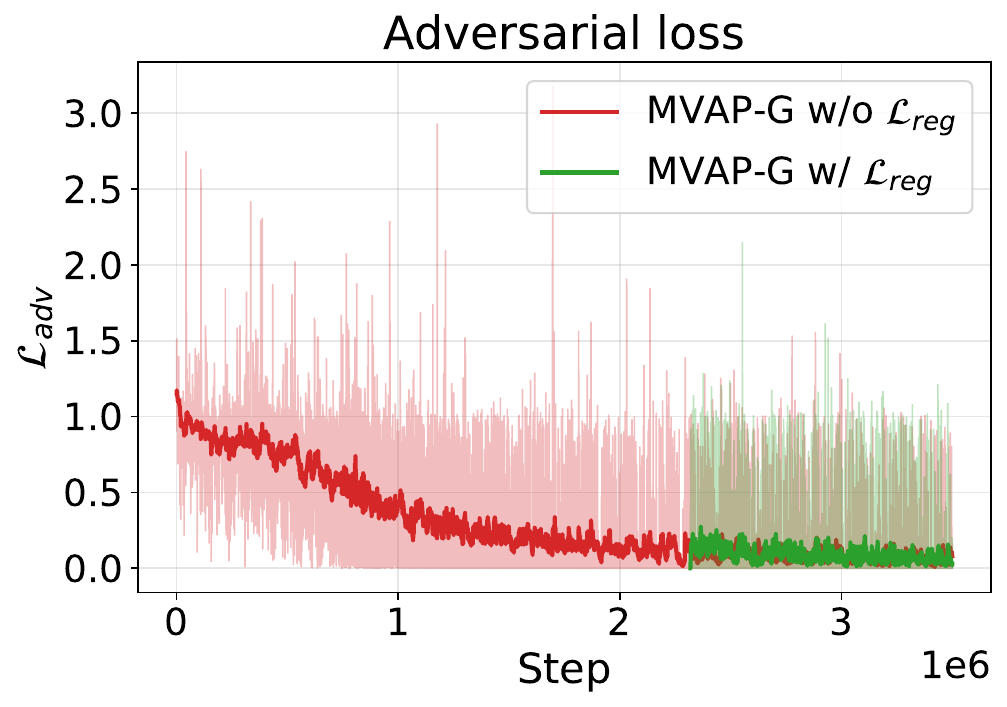}
    \hfill
    \includegraphics[width=0.32\linewidth]{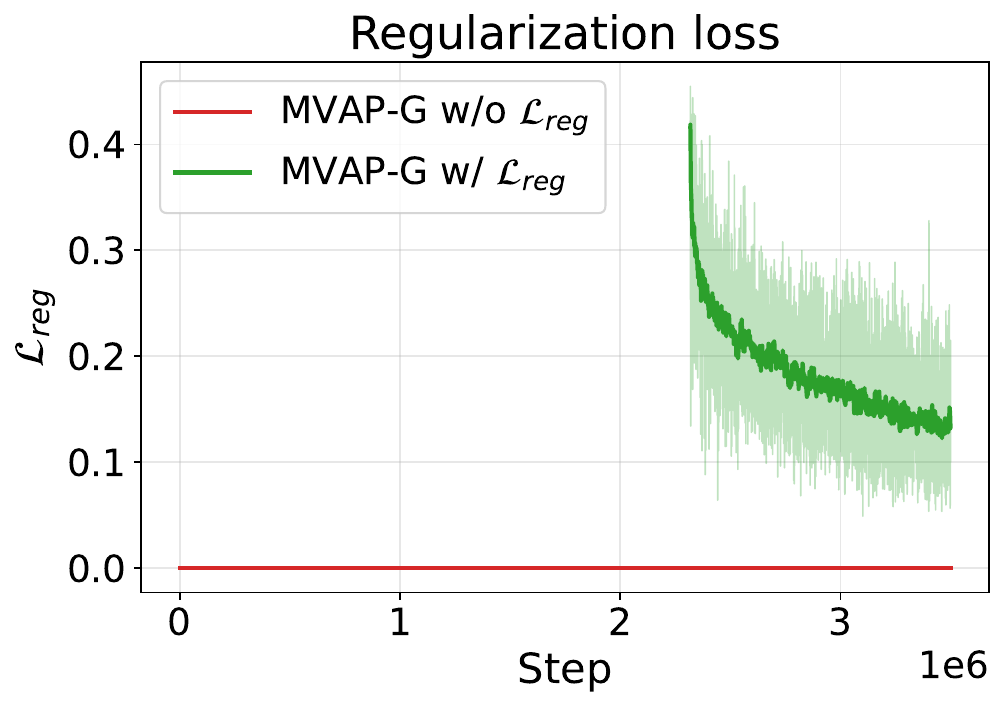}
    \vspace{-0.3cm}
    \caption{
    \textbf{(a) Cross-view adversarial alignment (CAA).} Attack strength remains steady even with more input views. Yet, w/o CAA faces inadequate attack performance against VGGT, especially in multi-view inputs.
    \textbf{(b) Adversarial loss.} The adversarial loss $\mathcal{L}_{\text{adv}}$ converges slowly with oscillations, aligning with the training process of generative adversarial networks~\cite{arjovsky2017wasserstein}.
    \textbf{(c) Regularization loss.} The convergence of $\mathcal{L}_{reg}$ indicates effective control over perturbation magnitude, ensuring visual imperceptibility while the adversarial loss remains low, thus achieving a balance between stealth and attack efficacy.
  }
  \label{fig:abl}
\end{figure*}

%% file: fig/exp-table-abl.tex
\begin{table}[!t]
\centering
\caption{Ablation study on key components of MVAP-G. Pre-training provides the largest boost in attack strength. The DPT outperforms the naive decoder, as it captures more features for pixel-level adversarial perturbations~\cite {ranftl2021vision}. CAA enables multi-view performance (more in Fig.~\ref{fig:abl} (a)). $\mathcal{L}_\text{reg}$ minimize the size of perturbation and ensures perturbation invisibility.}
\label{tab:component}
\resizebox{0.5\linewidth}{!}{
\begin{tabular}{lcccc}
\toprule
\multirow{2}{*}{Configuration} & \multicolumn{2}{c}{Attack performance} & \multicolumn{2}{c}{Invisibility} \\
\cmidrule(lr){2-3} \cmidrule(lr){4-5}
& CD $\uparrow$ & $\mathcal{L}_\text{adv} \downarrow$ & PSNR $\uparrow$ & LPIPS $\downarrow$ \\
\midrule
Baseline & 0.58 & 0.400 & 21.1 & 0.400 \\
+ Pre-training & 1.40 & 0.250 & 20.8 & 0.435 \\
+ DPT head & 1.68 & 0.120 & 20.5 & 0.478 \\
+ CAA & 1.75 & 0.008 & 20.3 & 0.512 \\
+ $\mathcal{L}_\text{reg}$ & 1.78 & 0.008 & 29.8 & 0.107 \\
\bottomrule
\end{tabular}
}
\end{table}

\begin{figure}[!b]
  \centering
  \vspace{-0.3cm}
  \includegraphics[width=0.70\linewidth]{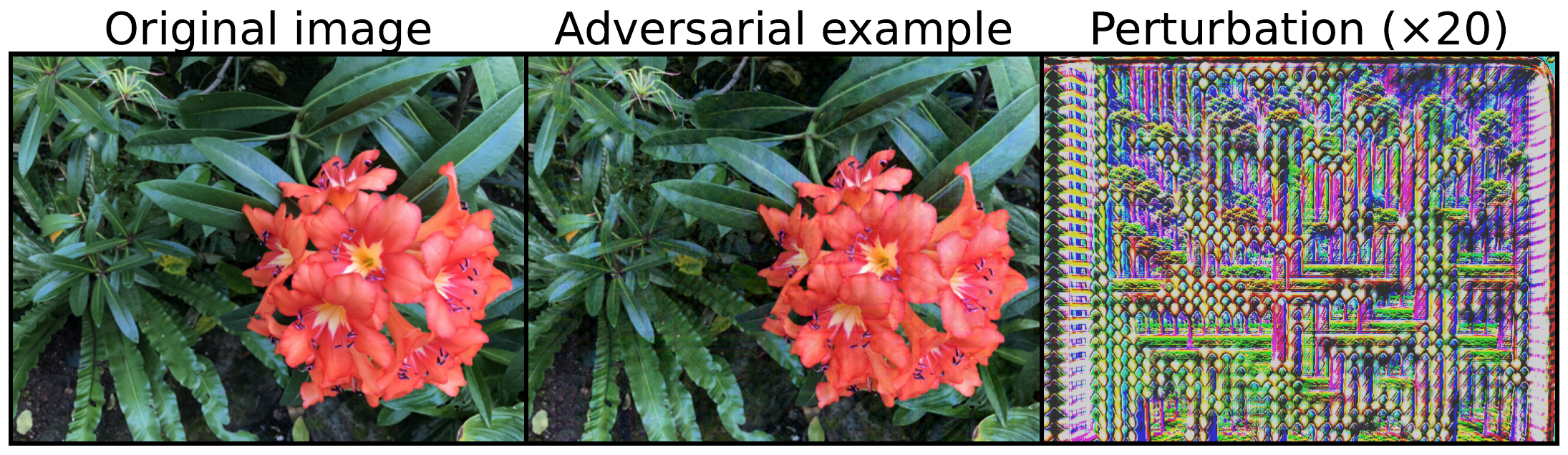}
  \caption{Adversarial examples and visualized $\times20$ perturbation.}
  \label{fig:adv_sample}
\end{figure}

%% file: fig/exp-adv-sample.tex
\begin{figure}[t]
    \centering
    \includegraphics[width=1.00\linewidth]{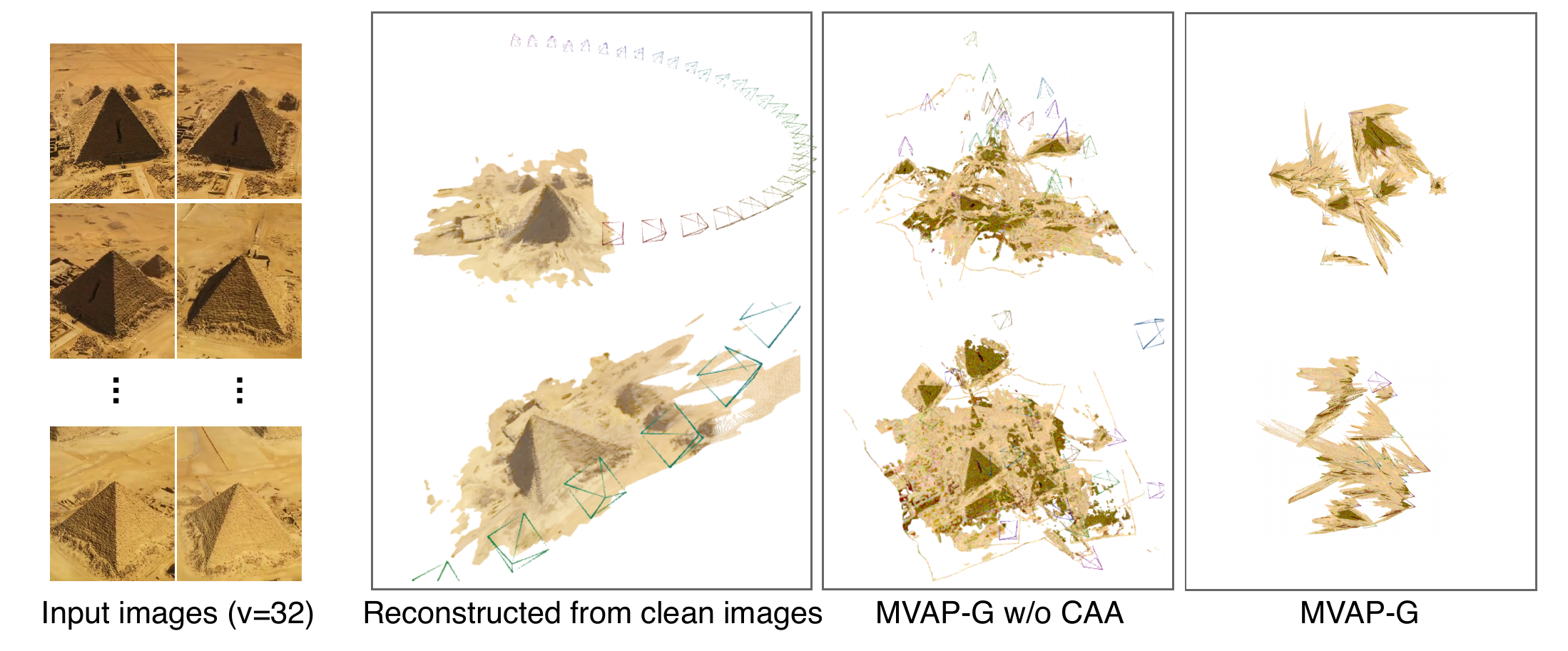}
    \vspace{-0.8cm}
    \caption{Impacts of CAA when input 32 view images. \textit{Without CAA}, perturbations allow VGGT to partially recover geometry results, yielding fragmented but recognizable structures.}
    \label{fig:caa}
\end{figure}

%% file: sec/5_conclusion.tex
\section{Conclusion}
In this work, we present the first systematic study of adversarial vulnerabilities in 3D foundation models, focusing on the Visual Geometry Grounded Transformer (VGGT). To address the limitations of traditional adversarial attack methods, we propose MVAP-G, a novel multi-view adversarial perturbation generator that produces multi-view-consistent adversarial perturbations in real time. Our method eliminates the need for per-scene optimization through a feed-forward generation pipeline, while explicitly maintaining adversarial performance across multi-view images. This work exposes critical security risks in 3D foundation models for adversarial robustness research.

\noindent \textbf{Limitations and future work.} Our method poses a serious threat to real-world VGGT deployment. This powerful attack could be misused to compromise safety-critical systems, such as autonomous driving or robotic navigation, causing severe real-world harm. 
Future work should prioritize the development of robust defenses and detection mechanisms, as well as responsible policy guidelines to prevent misuse.